\documentclass[runningheads]{llncs}
\usepackage[T1]{fontenc}
\usepackage{graphicx}
\usepackage{subcaption} 
\usepackage{caption}  
\usepackage{graphicx}
\usepackage{amsmath,amssymb,mathtools,mathrsfs}
\usepackage{booktabs,multirow}
\usepackage{algorithm2e}
\usepackage{url}
\usepackage{adjustbox}
\usepackage{marvosym}
\usepackage{rotating}
\usepackage{etoolbox}
\usepackage[pagebackref=true,breaklinks=true,letterpaper=true,colorlinks,bookmarks=false]{hyperref}
\usepackage{caption}

\BeforeBeginEnvironment{figure}{\vskip-2ex}
\AfterEndEnvironment{figure}{\vskip-2ex}
\begin{document}
\title{A Robust Unsupervised Domain Adaptation Framework for Medical Image Classification Using RKHS-MMD}
\titlerunning{RKHS-MMD: Unsupervised Domain Adaptation}
\author{Sapna Sachan\inst{1} \and 
Rakesh Kumar Sanodiya\inst{2} \and 
Amulya Kumar Mahto\inst{1}\Letter}
\authorrunning{Sachan et al.}
\institute{Mehta Family School of Data Science and Artificial Intelligence,\\
Indian Institute of Technology Guwahati, India\\
\email{s.sapna@iitg.ac.in, akmahto@iitg.ac.in}
\and
Indian Institute of Information Technology, Design and Manufacturing,\\
Jabalpur, India\\
\email{rakesh.pcs16@gmail.com}}
\maketitle              
\begin{abstract}%
Labeling medical images is a major bottleneck in the field of medical imaging, as it requires domain-specific expertise, and it gets further complicated due to variability across different medical centers and different imaging devices. Such heterogeneity introduces domain shifts and modality discrepancies, which limits the generalization of trained models. To address this important challenge, we propose an unsupervised Domain Adaptation framework that combines transfer learning with a Reproducing Kernel Hilbert Space–based Maximum Mean Discrepancy (RKHS-MMD) loss for the alignment of source and target domains. By jointly optimizing classification and RKHS-MMD losses, the methodology enhances generalization to unannotated medical datasets while diminishing reliance on manual annotation. Experimental evaluations presented on two chest X-ray datasets which are obtained from different medical centers shows outstanding improvements over models trained without adaptation. Furthermore, we perform a comparative study to see that RKHS-MMD performs better than the standard Maximum Mean Discrepancy (MMD) in reducing modality gap, emphasizing its effectiveness for medical image classification and also its strong capability in advance AI-driven medical diagnostics.

\end{abstract}%

\section{Introduction}
\begin{figure}[h!]
    \centering
    \begin{subfigure}[b]{0.3\textwidth}
        \centering
        \includegraphics[width=\linewidth,height=3cm]{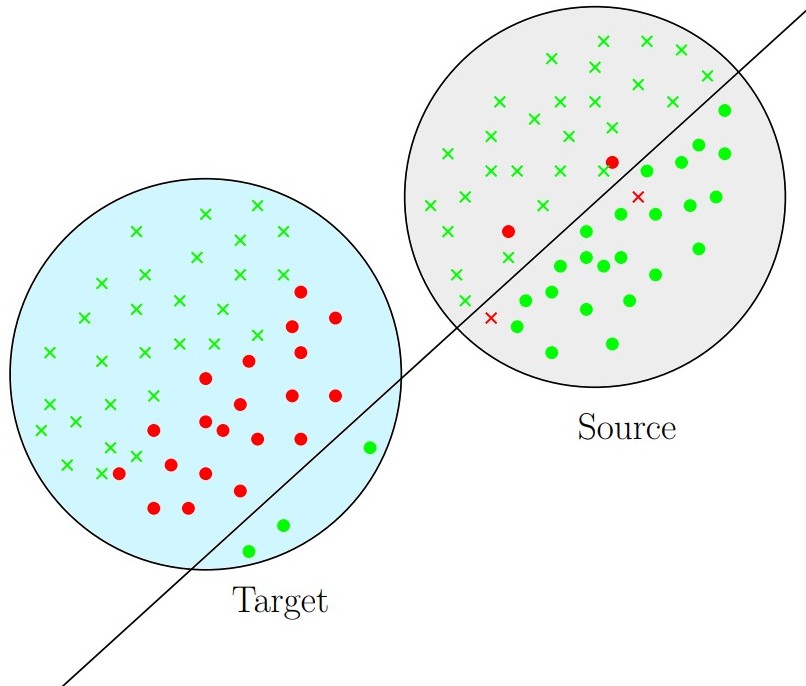}
        \caption{Initial Classification}
        \label{fig:a}
    \end{subfigure}
    \begin{subfigure}[b]{0.3\textwidth}
        \centering
        \includegraphics[width=\linewidth,height=3cm]{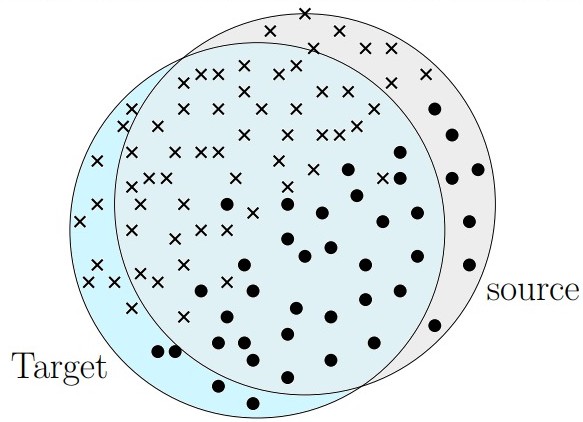}
        \caption{Domain Adaptation}
        \label{fig:b}
    \end{subfigure}
    \begin{subfigure}[b]{0.3\textwidth}
        \centering
        \includegraphics[width=\linewidth,height=3cm]{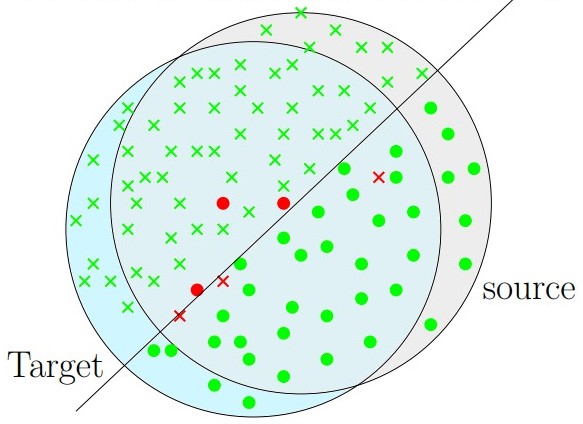}
        \caption{Final Classification}
        \label{fig:c}
    \end{subfigure}
    
    \caption{ An illustration of DA in classification. (a) Initial classification shows the mismatch between source and target domains, resulting in poor generalization. (b) Through DA, the feature distributions of the two domains are aligned. (c) Final classification demonstrates improved alignment and accurate predictions on the target domain.}
    \label{fig:three_images}
\end{figure}

Domain Adaptation (DA) is one of the widely used and studied methodologies for minimizing distributional gap between different datasets. There are many significant factors such as illumination, placed camera angle, aperture type, and distance of the object from the camera which play vital role in the appearance of images. These factors alter the appearance of image significantly, that is, they create differences in feature distributions making classification task challanging. As show in fig \ref{fig:three_images}, DA helps in bridging this gap and ensuring better alignment for enhanced classification.

Based on the availability of annotated data, DA may be classified into following two categories: $(i)$ Supervised Domain Adaptation (SDA) \cite{motiian2017unified}, and $(ii)$ Unsupervised Domain Adaptation (UDA) \cite{ganin2015unsupervised}. In SDA, the target dataset includes annotated data, whereas in the unsupervised approach, the target dataset is completely unannotated and such target domain data is prevalent in real-world scenarios. In medical AI applications, manual labeling of medical images remains one of the most significant challenges. Unlike natural images \cite{r4}, where annotation can often be automated or simplified, medical image labeling is labor-intensive and requires the expertise of physicians, radiologists, and other specialists. This process is, in general, time-consuming, tedious as well as costly. In addition, medical images are particularly sensitive to domain shift issues \cite{r3}, which arise because of variations in scanners, acquisition protocols, and patient cohorts. Such discrepancies often result in inconsistencies across different datasets and limits the ability of effective generalization of ML models. 

In the area of medical imaging, domain shift problem is very common, leading to an adverse affect on the model performance with even a subtle difference between datasets. In presence of these challenges, DA techniques \cite{r5} are used to reduce the differences which arise due to varying modalities and datasets. These techniques provide effective mechanism for knowledge transfer between different domains.  

The present study tackles the issue by utilizing the RKHS \cite{r1}, a high-dimensional mathematical framework that eases the alignment of feature distributions across domains. By embedding both annotated source data and unannotated target data into a RKHS, the differences in distributions between the modalities are reduced. To accomplish this, we develop a model utilizing annotated data from the source and unannotated data from the target, integrating a differentiable loss function that depends on MMD in addition to the classification loss. This combined optimization successfully identifies non-linear connections between domains, offering a strong approach to address modality gaps and domain shifts in medical imaging datasets.

\section{Related Work}

DA has been widely investigated to tackle the alterations in distributions between source and target domains, being a frequent obstacle in machine learning when the training (source) data differ from the testing (target) data. Initial shallow DA methods mostly depend on the alignment of statistical characteristics. For example, Subspace Alignment (SA) and Correlation Alignment (CORAL) \cite{r7} minimize differences by aligning feature subspaces or second-order statistics. Although these linear techniques are computationally efficient, they often struggle to represent the intricate non-linear patterns found in high-dimensional datasets.

With the rise of deep learning, research in DA began to focus on deep frameworks capable of learning representations that are invariant to domain changes. Deep CORAL \cite{r2} expanded upon CORAL for deep networks by integrating a differentiable covariance alignment loss, which allows for end-to-end optimization. Approaches that utilize adversarial learning, including Domain-Adversarial Neural Networks (DANN) \cite{r16}, utilized a gradient reversal layer to confuse a domain discriminator, thus promoting the extraction of domain-invariant features. Even though these techniques using adversarial learning are effective but they face some challenges such as training instability and mode collapse. On the other hand, methods that focus on discrepancies try to reduce the distributional variances. MMD \cite{r8,r9} based techniques make use of kernel embeddings for feature distributions alignment in high-dimensional spaces. For instance, in the Domain Discrepancy Criterion (DDC) \cite{r38}, MMD is integrated with AlexNet \cite{r15}, while Deep Adaptation Network (DAN) \cite{r40} advanced this concept of MMD by using multi-Kernel MMD (MK-MMD) across different layers to address problem with more complex domain discrepancies.Further, developments such as the Deep SubDA Network (DSAN) \cite{r41}, Domain Specific Adversarial Network (DSAAN) \cite{r42} and Deep Conditional Adaptation Network (DCAN) \cite{r44} brought profound enhancement in this area. These techniques significantly increaded robustness in different domains by incorporating class conditional alignment, auxiliary objectives and subdomain level matching.

In the field of medical imaging, domain shift is prevalent due to varying scanners, acquisition methods, and demographic differences among medical facilities \cite{r6} and these advancements have immense applications in medical imaging. Primary research used CORAL and Deep CORAL for harmonizing MRI data and for classification purposes for chest X-rays from different scan centers. But these linear approaches failed in handling complex , non-linear medical datasets appropriately. Kernel-based techniques like MMD and its variations emerged to more effectively align distributions in high-dimensional embedding spaces, resulting in improved adaptability for tasks such as lesion classification and organ segmentation. For instance, methods inspired by DSAN have shown enhanced performance in aligning cross-modal MRI and CT images by integrating conditional MMD losses. Additionally, adversarial techniques such as DANN \cite{r16}, Conditional Domain Adversarial Networks (CDAN) \cite{long2018conditional}, and Maximum Classifier Discrepancy (MCD) \cite{saito2018maximum} have become widely utilized in medical applications, covering areas from cross-center MRI brain tumor classification to robust diagnosis of chest X-rays. Furthermore, pseudo-labeling techniques have been applied in practical scenarios like COVID-19 detection \cite{n25}, where unannotated target data is successively annotated to enhance generalization. Recent developments, including Minimax Entropy (MME) \cite{saito2019semi} and Source Hypothesis Transfer (SHOT) \cite{liang2020we}, further improve robustness, especially when source data is unavailable during the adaptation process—a common situation in clinical settings due to privacy concerns. Together, these studies highlight the essential role of UDA in medical image classification, as it reduces modality discrepancies, boosts generalization across centers, and lessens the dependency on expensive manual annotations.

Even though these adversarial and discepancy based techniques have shown effective impact, conventional MMD- based methods face certain challenges, in particular, (1) The choice of kernels play important role. These methods are sensitive to selection of kernels as different kernels impact the capabilities of the methods in capturing non-linear discrepancies. 
Consequently, recent research has shifted toward RKHS formulations of MMD, which offer a more versatile embedding space for aligning intricate feature distributions. By concurrently optimizing classification and RKHS-MMD losses, models can enhance their generalization on unannotated medical datasets, surpassing traditional MMD-based approaches. Inspired by these developments, our research aims to merge RKHS-MMD with transfer learning for medical image classification. This method explicitly models feature alignment in RKHS, facilitating efficient adaptation across diverse domains without the necessity of target labels, thus promoting the use of AI-powered diagnostic systems.

\section{Methodology}
\begin{figure*}[h!]
\centering
  \includegraphics[width=\textwidth,height= 7cm]{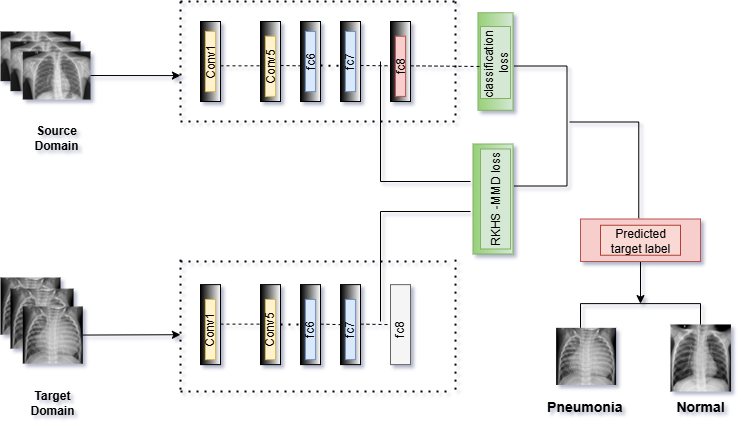}
    \caption{DA architecture using EfficientNetV2, combining classification and MMD losses to predict target labels from unannotated data.}
    \label{fig1}
\end{figure*}

This part provides a detailed description of our suggested methodology. As illustrated in fig \ref{fig1}, we leverage a CNN model (EfficientNetV2) as the foundational architecture for extracting features from the fully connected bottleneck layers in both source and target domains. In the unsupervised context, where annotated medical images from the source are available, we apply the conventional classification loss function – cross-entropy loss. Nevertheless, solely minimizing classification loss can cause overfitting to the source domain, which may diminish performance on the target domain. To improve generalization to the unsupervised target domain, we integrate a robust MMD loss within the RKHS, which helps ensure the extraction of domain-invariant features.

\subsection{Problem Definition}

The method jointly optimizes the classification loss and MMD loss:  
\[
L = L_{\text{CLASS}} + \sum_{i=1}^t \lambda_i L_{\text{MMD}},
\]
where \(t\) is the number of MMD loss layers in the deep network, and \(\lambda_i\) is a weight that balances DA with classification accuracy on the source domain. This joint optimization ensures that the two losses complement each other, reaching an equilibrium during training. At convergence, the final features are expected to be both well-aligned for DA and effective for target domain classification.

\subsection{Proposed Loss Function}
\subsubsection{Classification Loss}

The classification loss \(L_{\text{CLASS}}\) ensures accurate predictions on annotated source-domain data. Let \(D_S = \{(x_i, y_i)\}_{i=1}^{n_S}\) denote the source-domain training set, where \(x_i \in \mathbb{R}^d\) are input features and \(y_i \in \{1, 2, \ldots, C\}\) are class labels for \(C\) classes.The loss function is formulated as the cross-entropy between the predicted class probabilities and the true labels \cite{r20}.:

\[
L_{\text{CLASS}} = -\frac{1}{n_S} \sum_{i=1}^{n_S} \sum_{c=1}^C \mathbb{1}_{[y_i = c]} \log \hat{y}_{i,c},
\]

where:
\begin{itemize}
    \item \(\mathbb{1}_{[y_i = c]}\) is defined as indicator function which equals 1 if \(y_i = c\), and 0 otherwise,
    \item \(\hat{y}_{i,c}\) is the predicted probability of class \(c\) for sample \(i\), obtained via the softmax function applied to the logits \(z_i\).
\end{itemize}

\noindent The softmax function maps the logits \(z_i = [z_{i,1}, z_{i,2}, \ldots, z_{i,C}]\) to a probability distribution:
\[
\hat{y}_{i,c} = \text{softmax}(z_{i,c}) = \frac{\exp(z_{i,c})}{\sum_{j=1}^C \exp(z_{i,j})}.
\]

\noindent This formulation encourages the model to assign high probability to the correct class while penalizing incorrect predictions.

\subsubsection{RKHS and Kernel Foundations for MMD}
\paragraph{RKHS:}A \textbf{Hilbert space} is defined as a complete inner product space which implies that every Cauchy sequence convergeing to a limit within this space, also lies within the space. \textbf{Reproducing Kernel Hilbert Space}(RKHS) being a specific type of Hilbert space that has additional properties. 

Formally, let $\mathcal{H}$ be a Hilbert space of real-valued functions which is defined on a set $\Omega$. If there exists a function 
\[
k : \Omega \times \Omega \to \mathbb{R},
\] 
such that for any $x \in \Omega$, $k(\cdot, x) \in \mathcal{H}$, and $\forall$ $f \in \mathcal{H}$, the \textbf{reproducing property} holds:
\[
f(x) = \langle f, k(\cdot, x) \rangle_{\mathcal{H}},
\]
then $\mathcal{H}$ is an RKHS and $k$ is its \textbf{reproducing kernel}.  

The kernel function $k(x,y)$ must satisfy the following properties:  
\begin{enumerate}
    \item \textbf{Symmetry:} $k(x,y) = k(y,x)$,  
    \item \textbf{Positive definiteness:} For any finite set $\{x_1, \dots, x_N\} \subset \Omega$, the kernel matrix
    \[
    K = \big[k(x_i, x_j)\big]_{i,j=1}^N
    \]
    is positive semidefinite.  
\end{enumerate}

By the \textbf{Moore-Aronszajn theorem}, every positive definite kernel corresponds to a unique RKHS. This allows one to represent data in high-dimensional (potentially infinite-dimensional) feature spaces via kernel functions without explicitly computing the mapping. In this work, we employ the \textbf{Gaussian kernel}:
\[
k(x,y) = \exp\!\left(-\frac{\|x-y\|^2}{2\sigma^2}\right),
\]
which is widely used due to its smoothness and universal approximation properties.  

\paragraph{MMD Loss in RKHS:}  
The \textbf{MMD} is a distance metric in RKHS that measures how the distribution of the source domain is different from the distribution of the target domain. Let 
\[
D_S = \{x_i\}_{i=1}^{n_S} \subset \mathbb{R}^d
\] 
be the source-domain samples with labels $L_S = \{y_i\}_{i=1}^{n_S}$, and 
\[
D_T = \{u_i\}_{i=1}^{n_T} \subset \mathbb{R}^d
\] 
be the unannotated target-domain samples. The goal is to align the feature distributions of the source ($X_S$) and target ($X_T$) domains.  

The MMD distance in RKHS is defined as
\begin{align}\label{mmd}
    L_{\text{MMD}} = \|\mu_S - \mu_T\|_{\mathcal{H}}^2,
\end{align}
where $\mu_S$ and $\mu_T$ are the mean embeddings of the source and target distributions:
\[
\mu_S = \frac{1}{n_S} \sum_{i=1}^{n_S} k(x_i, \cdot), 
\quad
\mu_T = \frac{1}{n_T} \sum_{i=1}^{n_T} k(u_i, \cdot).
\]

\noindent The \textbf{empirical estimate} of $L_{\text{MMD}}$ is
\begin{align*}
    L_{\text{MMD}} &= \frac{1}{n_S^2} \sum_{i=1}^{n_S} \sum_{j=1}^{n_S} k(x_i, x_j) 
    + \frac{1}{n_T^2} \sum_{i=1}^{n_T} \sum_{j=1}^{n_T} k(u_i, u_j) \\
    &\quad - \frac{2}{n_S n_T} \sum_{i=1}^{n_S} \sum_{j=1}^{n_T} k(x_i, u_j).
\end{align*}

\noindent Minimizing $L_{\text{MMD}}$ aligns the $X_S$ and $X_T$ feature distributions, enabling \textbf{DA} in RKHS.”

\section{Experiment and Analysis}

\subsection{Description of Dataset}
\begin{figure}[h!]
    \centering
    \begin{subfigure}[b]{0.49\textwidth}
        \centering
        \includegraphics[width=\linewidth,height=3cm]{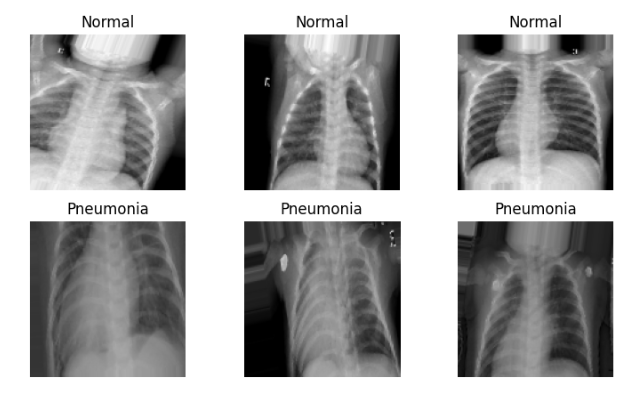}
        \caption{i.e.annotated Source Dataset}
        \label{fig a}
    \end{subfigure}
    \begin{subfigure}[b]{0.49\textwidth}
        \centering
        \includegraphics[width=\linewidth,height=3cm]{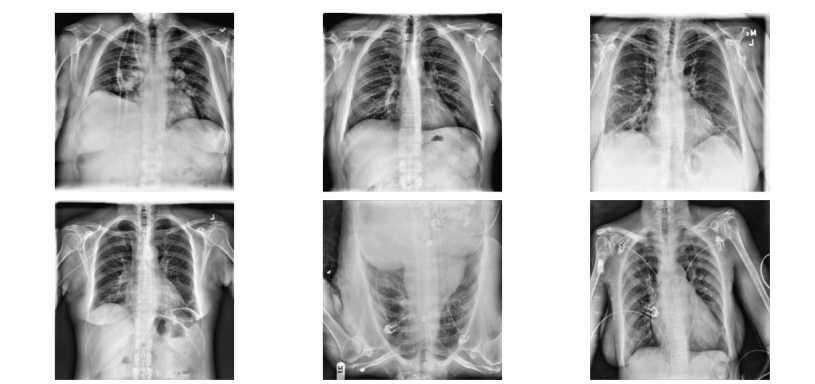}
        \caption{i.e.Unannotated target dataset}
        \label{fig b}
    \end{subfigure}
    \end{figure}

Table~\ref{table} summarizes the class-wise distribution of images across the training and testing datasets, while Fig.~\ref{fig a} and Fig.~\ref{fig b} illustrate representative samples from the annotated source and unannotated target domains, respectively. The table details the number of image samples for each class (Normal and Pneumonia) within both domains. In this setup, the training set consists of annotated source-domain data, whereas the testing set comprises unannotated target-domain data, aligning with the standard protocol for UDA.

\begin{table}[h!]
\footnotesize
\centering
\caption{ Data samples for Training and Testing}

\begin{tabular}{|c|c|c|c|}
\hline
\textbf{Set} & \textbf{Type} & \textbf{Source Domain} & \textbf{Target Domain} \\
\hline
\textbf{Training} & Normal & 1349 & - \\
\cline{2-4}
                  & Pneumonia & 3883 & - \\
\hline
\textbf{Testing}  & Normal & - & 3239 \\
\cline{2-4}
                  & Pneumonia & - & 1345 \\
\hline
\end{tabular}
\label{table}
\end{table}
\newpage The source dataset comprises 5,232 pediatric chest X-ray images \cite{r12}, categorized into two classes: Pneumonia and Normal. The target dataset includes 4,584 chest X-ray images from the study “Deep learning-based classification and referral of treatable human diseases” \cite{r13}. Although it also contains the same two classes, the images were collected from different medical centers, thereby emphasizing the domain shift between the source and target datasets.

\subsection{Experiments}\label{Experiments}

In this experiment, we use the EfficientNetV2 model \cite{r14} and follow the standard protocol for DA, utilizing all annotated source data and unannotated target data. The RKHS-MMD loss is applied to the final classification layer, with the last fully connected layer (fc) initialized with a normal distribution \( N(0, 0.005) \) and set to the number of categories. The learning rate for fc is set 10 times higher than for other layers, which are initialized with pre-trained ImageNet weights.

Training is done with a batch size of 32, 50-epochs, base learning rate of \( 1 \times 10^{-3} \), weight decay of \( 5 \times 10^{-4} \), and momentum of 0.9. The RKHS-MMD loss weight is tuned to balance the classification and MMD loss, ensuring the feature representation is discriminative while minimizing the domain gap between source and target data.

\subsection{Evaluation Metrics}
We utilize the Macro F1 Score as the evaluation metric. The Macro F1 Score is defined as the unweighted average of the F1 Scores for each class/label. The F1 Score for the \(i\)-th class is defined as:

\[
F1_i = \frac{2 \cdot \text{Precision}_i \cdot \text{Recall}_i}{\text{Precision}_i + \text{Recall}_i} 
\]

\noindent The Macro F1 Score is then computed as the unweighted average of the F1 Scores across all classes, such as for Normal and Pneumonia, which can be expressed as:

\[
\text{Macro F1 Score} = \frac{1}{C} \sum_{i=1}^{C} F1_i 
\]
\noindent where \( C \) representing the number of classes in total. The Macro F1 Score is particularly robust to data imbalance, ensuring that it remains a reliable performance metric even when the class distribution is skewed.

\subsection{Results and Discussion}\label{Results and Discussion}

We provide the outcomes for several models and comparison approaches utilized in this research. The effectiveness of the different models discussed in this paper is assessed using various metrics, including accuracy, classification loss, and RKHS-MMD loss.

\begin{table}[h]
\footnotesize
\centering
\caption{Comparison of models with Accuracy, Classification Loss, and RKHS-MMD Loss}
\begin{tabular}{|c|c|c|c|c|}
\hline
\textbf{Models} & \multicolumn{2}{c|}{\textbf{Accuracy}} & \textbf{Classification Loss} & \textbf{MMD Loss} \\
\hline
                & \textbf{Source Domain} & \textbf{Target Domain} &                              &                   \\
\textbf{EfficientNetV2}         & 1.00                   & 0.7745                   & 0.0014                         & 0.1092              \\
\textbf{EfficientNet}         & 0.8883                   & 0.6649                   & 0.1525                         & 0.1635              \\
\textbf{ResNet}         & 0.8096                   & 0.6132                  & 0.239                        & 0.0022              \\
\textbf{AlexNet}         & 0.6562                  & 0.514                   & 0.4417                        & 0.4734              \\
\textbf{VGG19}         & 0.6269                     & 0.5013                & 0.473                         & 0.3347             \\
\hline
\end{tabular}

\label{Table II}
\end{table}
In Table \ref{Table II}, we compare the performance of different models. Among them, EfficientNetV2 achieves the highest accuracy on both the source and target domains while maintaining the lowest classification loss, making it the most effective model for DA. This superior performance can be attributed to EfficientNetV2's compound scaling strategy, which balances network depth, width, and resolution more efficiently. As a result, the model learns richer feature representations that generalize better across domains, thereby reducing both classification and MMD losses.Furthermore, in table \ref{Table_III_CR}, the classification report for the target domain highlights the performance of the model for both "Normal" and "Pneumonia" classes using RKHS-MMD Loss.
 
\begin{table}[h!]
\footnotesize
\centering
\caption{Classification report for Target Domain}

\begin{tabular}{l|l|l|l|l|}
\cline{2-5}
                                   & \textbf{Precision} & \textbf{Recall} & \textbf{F1-Score} & \textbf{Support} \\ \hline
\multicolumn{1}{|l|}{\textbf{Normal}}       & 1.00      & 0.68   & 0.81     & 3239    \\ \hline
\multicolumn{1}{|l|}{\textbf{Pneumonia}}    & 0.56      & 1.00   & 0.72     & 1345    \\ \hline
\multicolumn{1}{|l|}{}             &           &        &          &         \\ \hline
\multicolumn{1}{|l|}{\textbf{Accuracy}}     &     -      &    -    & 0.77     & 4584    \\ \hline
\multicolumn{1}{|l|}{\textbf{Micro Avg.}}    & 0.78      & 0.84   & 0.77     & 4584    \\ \hline
\multicolumn{1}{|l|}{\textbf{Weighted Avg.}} & 0.87      & 0.77   & 0.78     & 4584    \\ \hline
\end{tabular}

\label{Table_III_CR}
\end{table}

\begin{figure}[h!]
    \centering
    \includegraphics[width=0.5\textwidth]{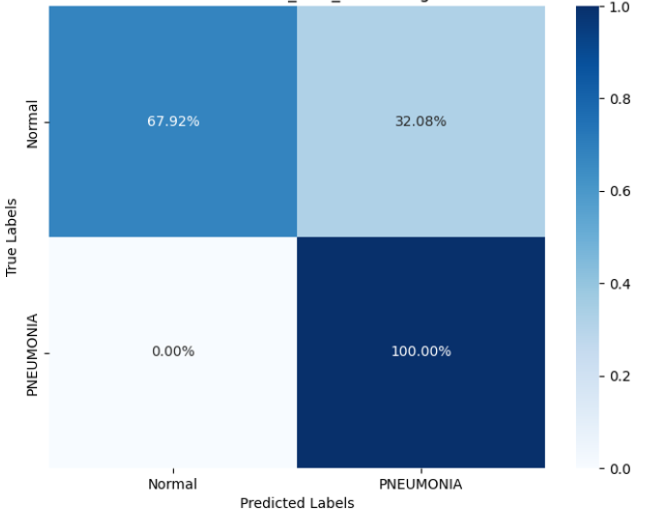}
    \caption{Confusion matrix}
    \label{fig 4}
\end{figure}

The confusion matrix in Fig. \ref{fig 4} visually represents the classification performance using RKHS-MMD showing correct and misclassified cases for "Normal" and "Pneumonia" classes.

\begin{table}[ht]
\footnotesize
\centering
\caption{Results with Deep RKHS-MMD, Standard MMD and Deep CORAL loss}
\begin{tabular}{|c|c|c|c|c|}
\hline
\textbf{Method} & \textbf{Accuracy} & \textbf{Classification Loss} &  \textbf{ Loss} \\ \hline
\textbf{RKHS-MMD } & 0.7745 & 0.0014  & 0.1092 \\ 
\textbf{Standard MMD} & 0.7201 & 0.0016  & 0.0470\\ 
\textbf{Deep CORAL} & 0.5488 & 0.0042 & 0.0991 \\ \hline
\end{tabular}
\label{tab:results}
\end{table}

\vspace{5mm}
 
\begin{figure}[h!]
    \centering
    \includegraphics[width=0.6\textwidth]{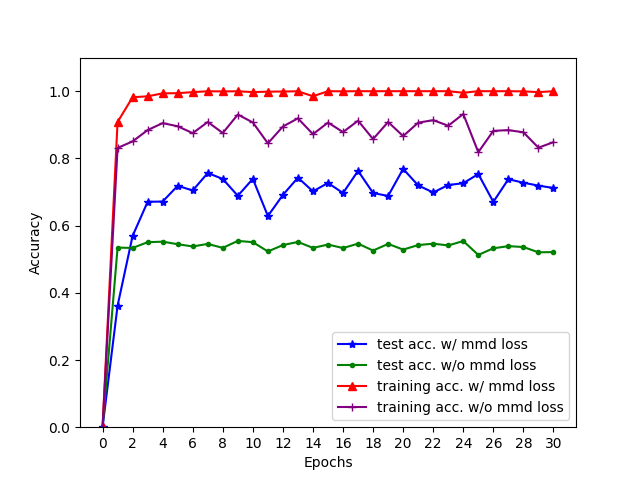}
    \caption{Comparison of DA performance using RKHS-MMD and without DA methods}
    \label{CoralMMD}
\end{figure}

 In table \ref{tab:results}, the superior performance of RKHS-MMD over Standard MMD and Deep CORAL is due to the way these methods handle distribution shifts. The Standard MMD loss used in deep DA typically measures the discrepancy between source and target distributions in a fixed kernel space, which often captures only limited aspects of the distribution (e.g., mean alignment with a specific kernel). This restricts its ability to model complex, higher-order discrepancies when the source and target domains differ significantly. In contrast, the RKHS-MMD formulation allows for a richer characterization of distribution differences by embedding the data into a RKHS(RKHS), where both mean and variance mismatches can be effectively captured. This provides a more robust and flexible alignment between source and target distributions, especially in scenarios involving non-linear shifts. On the other hand, Deep CORAL focuses on aligning only the second-order statistics (namely covariance) of the source and target features while ignoring higher-order moments. As a result, it becomes less effective when domain shifts involve more than covariance discrepancies.  In summary, RKHS-MMD outperforms both Standard MMD and Deep CORAL by leveraging the expressive power of RKHS to capture both mean and variance shifts, as well as complex non-linear relationships, thereby ensuring stronger domain alignment and improved target-domain performance. The graphs in Fig.\ref{CoralMMD} illustrate this performance gap, showing that the RKHS-MMD–based approach attains higher testing accuracy compared to models trained without any DA.

\section{Conclusion and Future Work}\label{conclusion}
In this work, we introduce an RKHS-based MMD loss for Unsupervised Domain Adaptation, targeting modality gaps in medical imaging. Our experiments show notable performance improvements, confirming the effectiveness of RKHS-MMD for end-to-end deep adaptation. The method integrates smoothly with standard architectures and addresses the challenge of limited annotated data. While evaluated on binary classification, it can be readily extended to multi-class settings, offering broad applicability in clinical practice.

{\small
\bibliographystyle{IEEEtran}
\bibliography{bibliography}

@book{r3,
  title={Dataset shift in machine learning},
  author={Qui{\~n}onero-Candela, Joaquin and Sugiyama, Masashi and Schwaighofer, Anton and Lawrence, Neil D},
  year={2022},
  publisher={Mit Press}
}

@inproceedings{motiian2017unified,
  title={Unified deep supervised domain adaptation and generalization},
  author={Motiian, Saeid and Piccirilli, Marco and Adjeroh, Donald A and Doretto, Gianfranco},
  booktitle={Proceedings of the IEEE international conference on computer vision},
  pages={5715--5725},
  year={2017}
}

@inproceedings{ganin2015unsupervised,
  title={Unsupervised domain adaptation by backpropagation},
  author={Ganin, Yaroslav and Lempitsky, Victor},
  booktitle={International conference on machine learning},
  pages={1180--1189},
  year={2015},
  organization={PMLR}
}

@article{r4,
  title={Imagenet large scale visual recognition challenge},
  author={Russakovsky, Olga and Deng, Jia and Su, Hao and Krause, Jonathan and Satheesh, Sanjeev and Ma, Sean and Huang, Zhiheng and Karpathy, Andrej and Khosla, Aditya and Bernstein, Michael and others},
  journal={International journal of computer vision},
  volume={115},
  pages={211--252},
  year={2015},
  publisher={Springer}
}

@article{r15,
  title={Imagenet classification with deep convolutional neural networks},
  author={Krizhevsky, Alex and Sutskever, Ilya and Hinton, Geoffrey E},
  journal={Advances in neural information processing systems},
  volume={25},
  year={2012}
}

@article{r1,
  title={Domain adaptive learning based on sample-dependent and learnable kernels},
  author={Lu, Xinlong and Ma, Zhengming and Lin, Yuanping},
  journal={arXiv preprint arXiv:2102.09340},
  year={2021}
}

@inproceedings{r2,
  title={Deep coral: Correlation alignment for deep domain adaptation},
  author={Sun, Baochen and Saenko, Kate},
  booktitle={Computer Vision--ECCV 2016 Workshops: Amsterdam, The Netherlands, October 8-10 and 15-16, 2016, Proceedings, Part III 14},
  pages={443--450},
  year={2016},
  organization={Springer}
}

@inproceedings{r5,
  title={U-net: Convolutional networks for biomedical image segmentation},
  author={Ronneberger, Olaf and Fischer, Philipp and Brox, Thomas},
  booktitle={Medical image computing and computer-assisted intervention--MICCAI 2015: 18th international conference, Munich, Germany, October 5-9, 2015, proceedings, part III 18},
  pages={234--241},
  year={2015},
  organization={Springer}
}

@article{r6,
  title={A survey of unsupervised deep domain adaptation},
  author={Wilson, Garrett and Cook, Diane J},
  journal={ACM Transactions on Intelligent Systems and Technology (TIST)},
  volume={11},
  number={5},
  pages={1--46},
  year={2020},
  publisher={ACM New York, NY, USA}
}

@inproceedings{r7,
  title={Return of frustratingly easy domain adaptation},
  author={Sun, Baochen and Feng, Jiashi and Saenko, Kate},
  booktitle={Proceedings of the AAAI conference on artificial intelligence},
  volume={30},
  number={1},
  year={2016}
}

@article{r8,
  title={Deep domain confusion: Maximizing for domain invariance},
  author={Tzeng, Eric and Hoffman, Judy and Zhang, Ning and Saenko, Kate and Darrell, Trevor},
  journal={arXiv preprint arXiv:1412.3474},
  year={2014}
}

@inproceedings{r9,
  title={Learning transferable features with deep adaptation networks},
  author={Long, Mingsheng and Cao, Yue and Wang, Jianmin and Jordan, Michael},
  booktitle={International conference on machine learning},
  pages={97--105},
  year={2015},
  organization={PMLR}
}

@article{r12,
  title={Labeled optical coherence tomography (oct) and chest x-ray images for classification},
  author={Kermany, Daniel},
  journal={Mendeley data},
  year={2018},
  publisher={mendeley}
}

@article{r13,
  title={Curated dataset for covid-19 posterior-anterior chest radiography images (x-rays)},
  author={Sait, Unais and Lal, K and Prajapati, S and Bhaumik, Rahul and Kumar, Tarun and Sanjana, S and Bhalla, Kriti},
  journal={Mendeley Data},
  volume={1},
  number={J},
  year={2020}
}

@inproceedings{r14,
  title={Efficientnetv2: Smaller models and faster training},
  author={Tan, Mingxing and Le, Quoc},
  booktitle={International conference on machine learning},
  pages={10096--10106},
  year={2021},
  organization={PMLR}
}

@inproceedings{r16,
  title={Exploring the limits of weakly supervised pretraining},
  author={Mahajan, Dhruv and Girshick, Ross and Ramanathan, Vignesh and He, Kaiming and Paluri, Manohar and Li, Yixuan and Bharambe, Ashwin and Van Der Maaten, Laurens},
  booktitle={Proceedings of the European conference on computer vision (ECCV)},
  pages={181--196},
  year={2018}
}

@article{r20,
  title={Generalized cross entropy loss for training deep neural networks with noisy labels},
  author={Zhang, Zhilu and Sabuncu, Mert},
  journal={Advances in neural information processing systems},
  volume={31},
  year={2018}
}

@article{r41,
  title={Deep subdomain adaptation network for image classification},
  author={Zhu, Yongchun and Zhuang, Fuzhen and Wang, Jindong and Ke, Guolin and Chen, Jingwu and Bian, Jiang and Xiong, Hui and He, Qing},
  journal={IEEE transactions on neural networks and learning systems},
  volume={32},
  number={4},
  pages={1713--1722},
  year={2020},
  publisher={IEEE}
}

@article{r42,
  title={Auc-oriented domain adaptation: from theory to algorithm},
  author={Yang, Zhiyong and Xu, Qianqian and Bao, Shilong and Wen, Peisong and He, Yuan and Cao, Xiaochun and Huang, Qingming},
  journal={IEEE Transactions on Pattern Analysis and Machine Intelligence},
  year={2023},
  publisher={IEEE}
}

@article{r38,
  title={Deep domain confusion: Maximizing for domain invariance},
  author={Tzeng, Eric and Hoffman, Judy and Zhang, Ning and Saenko, Kate and Darrell, Trevor},
  journal={arXiv preprint arXiv:1412.3474},
  year={2014}
}

@inproceedings{r40,
  title={Learning transferable features with deep adaptation networks},
  author={Long, Mingsheng and Cao, Yue and Wang, Jianmin and Jordan, Michael},
  booktitle={International conference on machine learning},
  pages={97--105},
  year={2015},
  organization={PMLR}
}

@article{r44,
  title={Unsupervised domain adaptation via deep conditional adaptation network},
  author={Ge, Pengfei and Ren, Chuan-Xian and Xu, Xiao-Lin and Yan, Hong},
  journal={Pattern Recognition},
  volume={134},
  pages={109088},
  year={2023},
  publisher={Elsevier}
}

@inproceedings{n25,
  title={Domain adaptation using pseudo labels for covid-19 detection},
  author={Yuan, Runtian and Li, Qingqiu and Hou, Junlin and Xu, Jilan and Zhang, Yuejie and Feng, Rui and Chen, Hao},
  booktitle={Proceedings of the IEEE/CVF Conference on Computer Vision and Pattern Recognition},
  pages={5141--5148},
  year={2024}
}

@article{long2018conditional,
  title={Conditional adversarial domain adaptation},
  author={Long, Mingsheng and Cao, Zhangjie and Wang, Jianmin and Jordan, Michael I},
  journal={Advances in neural information processing systems},
  volume={31},
  year={2018}
}

@inproceedings{saito2018maximum,
  title={Maximum classifier discrepancy for unsupervised domain adaptation},
  author={Saito, Kuniaki and Watanabe, Kohei and Ushiku, Yoshitaka and Harada, Tatsuya},
  booktitle={Proceedings of the IEEE conference on computer vision and pattern recognition},
  pages={3723--3732},
  year={2018}
}

@inproceedings{saito2019semi,
  title={Semi-supervised domain adaptation via minimax entropy},
  author={Saito, Kuniaki and Kim, Donghyun and Sclaroff, Stan and Darrell, Trevor and Saenko, Kate},
  booktitle={Proceedings of the IEEE/CVF international conference on computer vision},
  pages={8050--8058},
  year={2019}
}

@inproceedings{liang2020we,
  title={Do we really need to access the source data? source hypothesis transfer for unsupervised domain adaptation},
  author={Liang, Jian and Hu, Dapeng and Feng, Jiashi},
  booktitle={International conference on machine learning},
  pages={6028--6039},
  year={2020},
  organization={PMLR}
}
}

\end{document}